# Incremental Community Detection in Distributed Dynamic Graph

Tariq Abughofa
School of Computing
Queen's University
Kingston, ON, Canada
abughofa@queensu.ca

Ahmed A.Harby
School of Computing
Queen's University Kingston,
ON, Canada
Ahmed.harby@queensu.ca

Haruna Isah
School of Computing
Queen's University
Kingston, ON, Canada
h.isah@unb.ca

Farhana Zulkernine
School of Computing
Queen's University
Kingston, ON, Canada
Farhana.zulkernine@queensu.ca

*Abstract*— **Community detection is an important research topic in graph analytics that has a wide range of applications. A variety of static community detection algorithms and quality metrics were developed in the past few years. However, most real-world graphs are not static and often change over time. In the case of streaming data, communities in the associated graph need to be updated either continuously or whenever new data streams are added to the graph, which poses a much greater challenge in devising good community detection algorithms for maintaining dynamic graphs over streaming data. In this paper, we propose an incremental community detection algorithm for maintaining a dynamic graph over streaming data. The contributions of this study include (a) the implementation of a Distributed Weighted Community Clustering (DWCC) algorithm, (b) the design and implementation of a novel Incremental Distributed Weighted Community Clustering (IDWCC) algorithm, and (c) an experimental study to compare the performance of our IDWCC algorithm with the DWCC algorithm. We validate the functionality and efficiency of our framework in processing streaming data and performing large in-memory distributed dynamic graph analytics. The results demonstrate that our IDWCC algorithm performs up to three times faster than the DWCC algorithm for a similar accuracy.**

*Keywords—Distributed graph processing, dynamic graphs, streaming data, weighted community clustering*

## I. Introduction

Distributed processing of large-scale graphs has gained considerable attention in the last decade [1]. This is mainly due to the (i) unprecedented increase in the size of graph data such as the Web based social media networks, (ii) evolution of systems for processing massive graph data such as Pregel [2] and GraphX [3], and (iii) huge increase in the number of applications that utilize graph data such as traffic and social network analysis [4]. According to Heidari et al. [5], a typical graph processing system executes graph algorithms such as graph traversal over a graph dataset across five different logical phases, which include reading graph data, pre-processing, partitioning, computation, and error handling. Regardless of the size and type of framework or algorithms used, Heidari et al. reported [5] that large-scale graph data can be processed in offline, online, or real-time mode. Offline processing is the popular mode and is achieved by loading the graph dataset in memory from disk storage and processing it. Online processing allows users to update, maintain, and re-process the graph data automatically with new values either periodically or based on user-defined events. Real-time processing is similar to online processing except it also enables instant incremental updates to be made to the graph data. Thus, it requires the computation to be done immediately after changes happen to the data and the updated analytics results are returned with very short delays.

Several graph processing frameworks utilize static partitioning, which means that they consider the graph and the processing environment to remain unchanged [1][2][5]. However, most real-world graphs are dynamic as they change over time with new data producing new vertices and edges that need to be merged into existing graphs. The changes in dynamic graphs are further complicated by the need for real-time guarantees for applications such as real-time disease spreading and anomaly detection. Traditional static graph analytics approaches face a major limitation in meeting this demand [6]. Dynamic graph scenarios require novel online or real-time graph update and analytics algorithms since the traditional offline graph analytics approaches require first the whole graph to be updated with the new data, and then analytics algorithms to be applied to the whole graph, which is extremely computation-intensive and hence impractical.

Dynamic graph [7] updates can be node-grained or edge-grained. In node-grained dynamic graphs, new nodes or vertices are simultaneously added to the graph with all their incident edges. An example of such graphs is a network of scientific papers and their references. Once a paper is published, all the papers that it references are known as well and no new references (connections) are added later. In edge-grained dynamic graphs, new edges are added or removed for already existing vertices. Social networks are a good example of these graphs, people add new friends and "un-friend" old ones all the time. Thus, the assumption of knowing all the connections of a person when we add them to the graph is not viable. In these networks, the sequence of adding new edges is important and influences the evolution of the graph structure.

Recently, the problem of distributed processing of large dynamic graphs has gained considerable attention [7]. Several traditional graph operations such as the k-core decomposition [8-11], partitioning [12], and maximal clique computation [13] have been extended to support dynamic graphs. However, many graph processing frameworks do not support several graph operations in the context of dynamic graphs [4]. One such operation is community detection in graphs, which is the process of identifying groups of nodes that are highly connected among themselves and sparsely connected to the rest of the graph [14]. Such groups are referred to in the literature as "communities" and occur in various types of graphs. Several research studies on networks modeling real-world phenomena have shown that the networks are organized according to community structure and their structures evolve with time [15]. Therefore, community detection within large-scale graphs has become an important research problem [7][16][17]. It helps to discover new structural properties about the graph that cannot be found otherwise such as identification of the highly influential nodes known as community centroids [18]. It is also used for targeted marketing [14], distributed graph management [9][10], uncovering tightly connected entities in a graph [7], and finding major sub-graphs indicating special relationships that are generally obscured by the complex structure of the original graph [19].

Metrics for shaping communities often follow two approaches, either by maximizing the internal density of the communities by including heavily connected nodes into the community, or by reducing intra-community connectivity by removing weak connections among different communities [20]. Most of the existing community detection algorithms involve heavy computation and hence are time-consuming [21]. As the graphs being operated on become larger, the ability to process them in memory on a single machine becomes infeasible due to both time and memory constraints [14][17]. In dynamic graphs, the problem becomes more complex because the data keeps changing and the communities need to be adjusted by reapplying the solution to the whole graph every time the data changes [15]. With streaming data, communities need to be updated continuously or whenever a new micro-batch (too large of a batch size will lead to poor generalization, so micro-batches are needed to provide some basic intuition) of streaming data gets added to the graph. This poses a much greater challenge in devising a good community detection algorithm for dynamic graphs over streaming data.

In this paper, we propose an incremental community detection algorithm as a solution to the community detection problem for large dynamic graphs over streaming data. It gradually propagates new incoming data in the graph and adjusts the existing communities. The contributions of this study are as follows.

- We implemented the Distributed Weighted Community Clustering (DWCC) algorithm using Apache Spark [22] and GraphX [3][23] in Scala on a multi-cluster environment. The DWCC was proposed by Saltz et al. [14] which was implemented on the Pregel platform for static data.
- We conducted an extensive performance study of the DWCC algorithm to identify the costly operations to optimize the processing time and memory consumption.
- Based on the results of the above study, we developed a novel Incremental Distributed WCC (IDWCC) algorithm for undirected and unweighted node-grained dynamic distributed graphs. IDWCC applies the Weighted Community Clustering (WCC) optimization technique to add new vertices from the streaming data to the most suitable communities in an existing distributed graph. We implemented the algorithm in Scala using GraphX to work with Spark Streaming. To the best of our knowledge, this is the first node-grained incremental distributed community detection algorithm.
- We experimentally validated both DWCC and IDWCC algorithms and compared their performances using real-world datasets with ground-truth communities. The evaluation addresses the performance, quality, and applicability aspects.

The remainder of this paper is organized as follows. We outline the existing solutions for the community detection problem and explain the WCC metric in Section II. In Section III we describe our implementation of the DWCC algorithm using Spark and GraphX and the propose the IDWCC algorithm. Next, we present a complexity analysis and experimental evaluation of the two algorithms in Section IV. Section V presents a case study of WCC optimization in dynamic graphs for product recommendations. Finally, we conclude this study and outline further improvements in Section VI.

## II. BACKGROUND & TERMINOLOGY

Community detection is a widely studied problem [24]. It is one of the most relevant topics in the field of graph data processing due to its importance in many fields such as biology, social networks, or network traffic analysis [20]. In this section, we present a brief literature review of some of the work in this area and explain the key concepts behind the WCC optimization technique.

### A. Literature Review

A variety of community detection algorithms have been developed based on different graph update strategies during the past few years. Label Propagation [25][26] is one of the most popular community detection methods, which is implemented in GraphX [3]. This algorithm chooses the community of the current node using the labels of its neighboring nodes. Initially, each node is initialized with a unique label and at every iteration of the algorithm, each node adopts the label that most of its neighbors have. As the labels propagate through the network, densely connected groups of nodes form a consensus on their labels. At the end of the algorithm, nodes having the same labels are grouped as communities. Another popular community detection method based on random walks is Infomap [27]. Finding community structure in networks using Infomap is equivalent to solving an information flow problem. Rosvall and Bergstrom [27], exemplified this by making a map of science, based on how information flows among scientific journals through citations. A detailed survey and guided tour through the main aspects of community detection methods and their applications have been outlined by Harenberg et al. [28] and Fortunato et al. [29] respectively.

Many centralized community-detection methods have been proposed in the literature, however, recent dramatic growth in real-world network size requires community detection to be performed in a distributed environment [30]. Apart from the huge sizes, modern networks are characterized by high dynamics, which challenges the efficiency of community detection algorithms [31]. These challenges have led to several research solutions on distributed community detection in both static and dynamic graphs. Hung et al. [32] modeled community detection on edge-labeled graphs as a tensor decomposition problem and proposed a fast, accurate, and scalable distributed system for community detection in large static graphs based on the Spark framework. Clementi et al. [30] introduced a dynamic community detection framework that relies on the Label Propagation algorithm [26][27]. However, the framework was evaluated using randomly generated networks rather than real-world graphs.

Recently, Jian et al. [31] designed an algorithm based on the Label Propagation method [26][27] that can incrementally detect communities over distributed and dynamic graphs. According to Jian et al., besides detecting high-quality communities, the algorithm can incrementally update the detected communities after a batch of edge insertion and deletion operations. The algorithm was implemented by using the MapReduce model. The evaluation results on real-world datasets show that the algorithm can detect communities incrementally with a running time that is sublinear to the changed edge number. What is not clear, however, in the evaluation is the measure of the indicator of the quality of the communities for a real-world dataset.

Several metrics such as modularity and conductance have been proposed as indicators of the quality of a community in a graph [19]. Modularity is considered the most prominent quality measure for community detection [24][33]. It prioritizes communities based on their internal edge density. One of the most popular algorithms based on modularity optimization is the Louvain algorithm, which is presented in detail by Blondel et al. [33]. This algorithm is a greedy optimization that can be used for weighted graphs. The algorithm starts with each vertex as its own community. Then it progresses in an iterative manner where each iteration consists of two phases. The first phase calculates the gain in modularity (see Eq. 1) by adding each vertex to a neighboring community and to a community that produces the highest gain.

$$\Delta Q = \left[\frac{\Sigma in + k_{i,in}}{2m} - \left(\frac{\Sigma tot + k_i}{2m}\right)^2\right] - \left[\frac{\Sigma in}{2m} - \left(\frac{\Sigma tot}{2m}\right)^2 - \left(\frac{k_i}{2m}\right)^2\right] \quad (1)$$

This gain in modularity $\Delta Q$ when a node is moved into a community $C$ is calculated using Eq. 1. $\Sigma in$ is the sum of the weights of the links inside $C$, $\Sigma tot$ is the sum of the weights of the links incident to nodes in $C$, $k_i$ is the sum of the weights of the links incident to node $i$, $k_{i,in}$ is the sum of the weights of the links from i to nodes in C, and m is the sum of the weights of all the links in the network.

More recently another metric called the Weighted Community Clustering (WCC) was introduced by Prat-Pérez et al. [20] to evaluate the quality of communities based on their density in terms of triangles. Unlike Louvain, WCC optimization does not consider edge weights in the computations. The WCC metric ensures that communities are cohesive, structured, and well defined. It is used in the Scalable Community Detection (SCD) algorithm [17] for detecting communities in undirected unweighted graphs of unprecedented size in a short execution time. A distributed version of the algorithm based on the vertex-centric paradigm was developed later by Saltz et al. [14] on the Pregel platform [2]. This approach performs well on static graphs of over one billion edges. However, most real-world graphs are not static but often change over time. The changes are usually represented as streaming networks where data need to be added to a network incrementally in real-time while updating the graph community structure [7]. Therefore, a solution is needed to add new data and update communities in distributed dynamic graphs in a multi-cluster environment for streaming data.

For incremental community detection, many modularity-based solutions have been proposed but very few solutions exist for node-grained graphs. Shang et al. [36] introduced an algorithm that depends on the Louvain algorithm for detecting an initial community structure as well as the communities for new vertices. Pan et al. [37] developed a method for edge-grained graphs. The problem with this method is that it assumes the edges are added in a certain order. As a result, it cannot handle node-grained graphs properly where the edges are added simultaneously, and gives poor performance [7]. A recent method called the Node-Grained Incremental (NGI) community detection based on modularity optimization was proposed by Yin et al. [7] for node-grained dynamic graphs. However, it was only implemented for centralized but not distributed processing.

In this paper, we propose an incremental community detection algorithm for large distributed dynamic graphs on a multi-cluster environment based on the WCC optimization technique. The WCC optimization algorithm is explained in detail by Prat-Perez et al. [14][17]. In this section, we summarize the fundamental concepts of the WCC metric, its applications in community detection in large graphs, and the processing steps namely pre-processing and partitioning.

*B. WCC*

Prat-Pérez et al. [20] [19] first introduced the metric called Weighted Community Clustering (WCC) to evaluate the quality of community partitioning based on the distribution of triangles in the graph. The WCC optimization approach constructs triangles of vertices in the graph to measure the density of vertices. WCC optimization has gained a lot of attention due to less computational complexity as it does not consider edge weights in the computations and demonstrates superior results over other commonly used metrics like modularity [17].

Given a graph $G(V, E)$ composed of a set of vertices $V$ and a set of edges $E$, $t(x, V)$ denotes the number of triangles that pass through the vertex $x$ and links it to neighbouring vertices in a set of $V$ vertices. (triangle count for $x$), and $vt(x, V)$ denotes the number of neighboring vertices that close at least one triangle with $x$ for each vertex in the graph. Given a community $C$ in graph $G$, $t(x, C)$ and $vt(x, V)$ are the same as the previous measurements considering the vertices inside $C$ only. Based on these four measurements, the WCC value for a vertex $x$ in a community $C$ can be calculated using Eq. 2 as explained by Prat-Pérez et al [17].

$$WCC(x, C) = \begin{cases} \frac{t(x, C)}{t(x, V)} \cdot \frac{vt(x, V)}{|C\{x\}| + vt(x, V\setminus C)} & \text{if } t(x, V) \neq 0 \\ 0 & \text{otherwise} \end{cases} \quad (2)$$

The WCC value for the whole graph is calculated from the average of the WCC of all the vertices in all the communities in the graph as described in Eq. 3.

$$WCC(G) = \frac{1}{|V|} \sum_{i=1}^{n} WCC(x, C_i) \quad (3)$$

Prat-Pérez et al. [20] introduced a set of basic properties that any community cohesion metric for social networks should fulfill. These properties include (i) clustering coefficient, defined as the probability that two neighbors of a given individual are also neighbors themselves [24], (ii) the dynamics of community formation, (iii) presence of a bridge, an edge which if removed from the graph, creates two separate connected components, (iv) presence of a cut vertex, a node whose removal splits the graph into two or more connected components, and (v) presence of clique, a vertex connected to another vertex with an edge which forms a maximal clique. The authors further proved that WCC is a good candidate to distinguish communities in social networks. In terms of the clustering coefficient, they discovered that WCC reacts to the internal structure of the communities, and in particular, to the presence of triangles. Regarding the appearance of a new node in a community, WCC was found to have a better value for a node with fewer connections if the node was included in the community. It has, however, a better value for a node with many connections if the node was kept outside the community. They also discovered that WCC was resistant to bridges, and an optimal community in social networks can not contain a bridge. Finally, WCC was found to be able to separate communities into two cliques.

As stated before, several metrics such as modularity and conductance have been proposed as indicators of the quality of a community in a graph. However, we chose the WCC metric and its optimization method to be the basis of our distributed dynamic graph community detection algorithm because of its performance, increasing popularity in the graph processing community, and potential in ensuring that communities are cohesive, structured, and well-defined [20]. WCC provides a good trade-off between performance and quality [14][16][17]. In addition, the optimization process of WCC can be distributed easily; the calculations of the best movement and the WCC value for each vertex can be done locally, and thus the computations can be executed in parallel. To the best of our knowledge, it is the most efficient solution for community detection in large-scale graphs.

## III. System Design

WCC is used in the SCD algorithm [17] for community detection in centralized graphs. A distributed version of the algorithm exists for static distributed graphs, which was implemented by Saltz et al. [14] in Java for the Graph processing engine. In this paper, we propose an Incremental Distributed Weighted Community Clustering (IDWCC) algorithm for detecting communities incrementally in a distributed dynamic graph that is continuously updated from streaming data. Communities help in clustering very large graph data on a distributed infrastructure for better management and fast processing of analytical queries.

We validate the algorithm using our existing multi-level streaming data processing framework. The framework uses Spark, GraphX, and GraphFrames to create and maintain a dynamic distributed graph. Since the implementation of the community detection algorithm based on WCC using these tools did not exist, we implemented one using Scala, GraphX, and GraphFrames for distributed processing on Spark.

We describe the three basic steps of the WCC optimization algorithm. Then we illustrate the Spark implementation of the Distributed WCC (DWCC) algorithm for a static distributed graph. Finally, we explain and demonstrate our IDWCC for detecting communities incrementally in dynamic distributed graphs.

### A. Partitioning

In this step, we compute an initial partition of the graph. First, the vertices are sorted by their clustering coefficients in descending order. Then the vertices are iterated on and for each vertex $x$ not previously visited, we create a new community $C$ that contains $x$ and all its neighbors that were not visited before. The algorithm requires the following conditions to be met in an initial partition.

- Every community should contain a single-center vertex and a set of border vertices connected to the center vertex.
- The center vertex should be the vertex with the highest clustering coefficient in the community.
- Given a center vertex $x$ and a border vertex $y$ in a community, the clustering coefficient of $x$ must be higher than the clustering coefficient of any neighbor $z$ of $y$ that is the center of its own community.

In the final step, the initial partition is improved iteratively using a hill-climbing method. The execution stops when no further improvements to the global WCC can be achieved, or when a predefined number of iterations do not provide any significant improvement as specified by a threshold. Next, we will discuss our distributed implementation of WCC optimization for GraphX. The proposed IDWCC algorithm is explained after that.

The Pregel API in GraphX helps in executing the partitioning of a distributed graph while respecting all the initial partitioning conditions. It performs an iterative execution process in which dynamic vertices keep broadcasting changes in their communities to their neighbors while receivers update their communities depending on the change notifications, they receive from the neighbors until no further adjustments are needed.

Computing the improvement of the global WCC using Eq. 3 requires the computation of the internal triangles of each community of the graph, which makes it inefficient to compute all possible movements of each vertex. Prat-Perez et al. [17] present a heuristic for calculating WCC improvement caused by moving a single vertex to a new community using the statistics about the vertex and its neighboring communities. The heuristic as presented in Eq. 4, gives an approximated value and does not require the computation of the internal triangles of each community. Instead, it depends on calculating the following statistics: $d_{in}$: the number of edges that connect the vertex $v$ to the vertices inside the community $C$ where it is moving, $d_{out}$: the number of edges that connect $v$ to the vertices outside $C$, $b$: the number of edges that are in the boundary of $C$, $\delta$: the edge density of $C$, $r$: the number of vertices in $C$, and $w$: the clustering coefficient of the graph. We use the same heuristic due to its efficiency. Since this computation occurs independently within each vertex, all vertices may perform their movements simultaneously, meaning that this part of the algorithm can be distributed effectively on multiple compute nodes to be executed in parallel to improve the performance of the algorithm.

$$WCC'_l(x, C) = \frac{1}{|V|} \cdot (d_{in} \cdot \Theta_1 + (r - d_{in}) \cdot \Theta_2 + \Theta_3) \quad (4)$$

Prat-Pérez et al [17] described $\Theta_1$, $\Theta_2$, and $\Theta_3$ as the WCC improvements of the vertices in $C$ that are connected to $x$, the vertices in $C$ that are not connected to $x$, and the vertices $v$ respectively, where $v$ represents the set of vertices to be added to community $C$.

B. *Optimization*

We implement DWCC optimization for Apache Spark using its distributed in-memory graph structure, GraphX. The implementation is somewhat influenced by the existing graph processing libraries in Spark and the properties of the GraphX structure. We calculated the execution time for each small step of DWCC as shown in Figure 2. Based on these calculations, we developed an algorithm that works in three phases. First, it merges the batch with the maintained evolving graph, updates the vertex statistics, and optimizes the graph. Second, it assigns the new vertices to initial communities. Finally, it optimizes the WCC metric to generate better communities.

As a first step, a new graph $G^* = (V^*, E^*)$ is generated from the newly arrived batch $\delta^*$. The produced graph is then merged with the full graph to produce $G_{t+1} = (V_t \cup V^*, E_t \cup E^*)$ as demonstrated in Fig. 1.

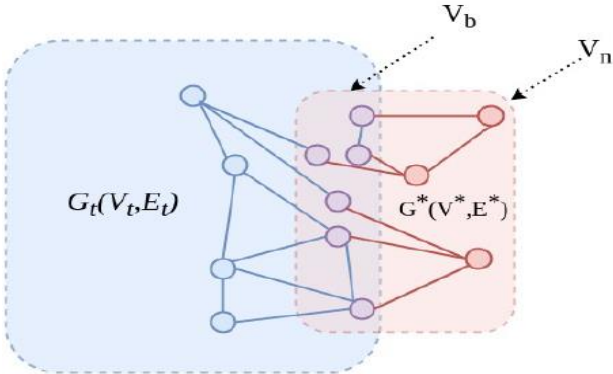

*Figure 1. Merging $G^*$ with $G_t$.*

We identify a set of vertices which we call the *border vertices*. These vertices exist in both $G_t$ and $G$, are a part of the edges that connect the newly arriving batch with the old graph. Let us denote this set as $V_b = V_t \cap V^*$. We refer to the rest of the vertices in the new graph which are not part of the border vertices, as the *inner vertices* $V_n = V^* \setminus V_b$. The problem with the border vertices is that they have already been assigned to communities in $G_t$. But since they have new connections, they are likely to belong to different communities. We isolate each of these vertices in its own community in the full graph $G_{t+1}$. The merge phase also calculates $t(x, V_{t+1})$ and $vt(x, V_{t+1})$ for each vertex $x$ in $G_{t+1}$. To perform the calculations efficiently, we recognize three possible situations. (a) Statistics of the old vertices stay the same as they were for the previous micro-batch $t$. (b) The inner vertices need to calculate the statistics. (c) The border vertices have new connections and thus might belong to new triangles and need to update their statistics. The definition of the stream batches, which is presented in Eq. 7, is important for updating the statistics of the border vertices as it assures that the graph holds the following conditions. Let's denote the set of triangles that pass through a vertex $x$ in graph $G$ as $T_{x,G}$ and the set of vertices that form at least one triangle with $x$ as $VT_{x,G}$. Then the following holds true.

$$T_{X,G_t} \cap T_{x,G^*} = \emptyset \text{ and } VT_{x,G_t} \cap VT_{x,G^*} = A$$

where $A$ is the set of vertices that are neighbors of $x$ and form triangles with it in both $G_t$ and $G^*$. Based on these statements, the statistics for the border vertices are calculated as follows.

$$t(x, G_{t+1}) = t(x, G_t) + t(x, G^*) \quad (5)$$

$$vt(x, G_{t+1}) = vt(x, G_t) + vt(x, G^*) - |A| \quad (6)$$

Using these two measurements we can compute the local clustering coefficient for each vertex and the global clustering coefficient $w$ which is needed to calculate $WCC'_l$. At the end of this phase, we optimize the graph in the same way as it is done for DWCC to reduce the memory consumption and the processing required in the succeeding phases, which is a relatively cheap operation (see Fig. 2).

| **Algorithm 1:** Partitioning |
|---|
| 1:     Let $P$ be a set of communities generated at the last micro-batch; |
| 2:     $S \leftarrow$ sort *ByClusteringCoefficients($V_{t+1}$)*; |
| 3:     **for all** $v$ in $S$ **do** |
| 4:         **if** *notVisited(v)* **then** |
| 5:             *markAsVisited(v)*; |
| 6:             **if**   $v \in V^*$ **then** |
| 7:                 $C \leftarrow \{v\}$; |
| 8:             **else** |
| 9:                 $C \leftarrow P.getCommunity(v)$; |
| 10:            **for all** $u$ *in neighbors(v)* **do** |
| 11:                **if**  *notVisited(u)* **then** |
| 12:                    *markAsVisited(u)*; |
| 13:                    **if** $u \in V^*$ **then** |
| 14:                        *C.add(u)*; |
| 15:        $P.add(C)$ |

We choose communities for the vertices that appear in the new batch. These vertices include the inner vertices $V_n$ which have no communities assigned to them yet, and the border vertices $V_b$ which were removed from their communities during the previous phase. We use the same algorithm as used in DWCC (see Algorithm 1), but we limit it to the above-mentioned sets of vertices only. Hence, every vertex in the new batch chooses the vertex with the highest clustering coefficient that does not belong to a community of another vertex as its community center.

Algorithm 2 follows the same steps as its counterpart Algorithm 1, the DWCC algorithm. However, it includes two

optimizations since it is the most expensive processing phase in terms of computations:
- Calculation of the community movements is still done on all the vertices, but we drop calculating the value of WCC in each iteration.
- We use a fixed number of iterations rather than using more iterations when good WCC improvement appears.

This might result in missing community movements that can have a good impact on WCC. However, as we process subsequent micro-batches, all the vertices start changing their communities again and any previous changes that were missed are subsequently recovered. This way, the degradation of WCC over time is avoided.

**Algorithm 2:** Partitioning optimization
1: Let $P$ be the initial partition;
2: $iteration \leftarrow 1$;
3: **Repeat**
4:    $M \leftarrow \emptyset$
5:    **For all** $v$ in $V$ **do**
6:       $M.add(bestMovement(v, P))$
7:    $P \leftarrow applyMovements(M, P)$;
8:    $Iteration = iteration+1$;
9: **until** $iteration > maxIterations$;

*C. Preprocessing*

This phase aims to calculate the $t(x, V)$ and $vt(x, V)$ values for each vertex of the graph. After these measurements are calculated, a graph optimization which is stated in the optimization section, is performed by removing edges that do not close any triangles.

The Triangle Count algorithm in GraphX[1] requires the graph to be canonical which means that the graph should ensure the following:
- Free from self-edges (edges with the same vertex as a source and a destination).
- All its edges are oriented (the source vertex has a greater number of directly connected triangles than the destination vertex based on a pre-defined comparison method).
- Has no duplicate edges.

The cleaning is done using the subgraph API provided by GraphX. We keep the calculated statistics namely, the triangle count and the degree of vertex for later use. We took advantage of the fact that GraphX supports property graphs and hence we can save these statistics as properties of the graph vertices.

*D. Implementation*

We modify certain steps of the DWCC algorithm which incur high computational cost to make the algorithm more scalable so that we can apply it to distributed dynamic graphs. The key steps of the DWCC algorithm are as follows.
- Step 1: Vertex Statistics (preprocessing): Count the triangles of vertices to identify communities and keep the statistics of triangle count and the degrees of vertices.
- Step 2: Graph Restructuring (preprocessing): Move vertices and remove redundant edges as a part of graph restructuring, optimizing, and cleaning.
- Step 3: Iterative Partitioning (initial partitioning): Use broadcasting protocol for dynamic vertices to add/modify communities on distributed graphs.
- Step 4: WCC Optimization (partition refinement): Compute the WCC metric or improvement for the graph.

We investigate and calculate the execution time of each of the small steps of DWCC and propose changes to the DWCC algorithm to define the IDWCC for node-grained dynamic distributed graphs. The IDWCC algorithm has many similar steps as the DWCC. However, it has optimizations that avoid repeated calculations and consequently reduce the memory, data movement, and computational costs without sacrificing the quality of the result.

In this paper, we limit our scope of interest to dynamic graphs that satisfy two properties. First, the graph progresses over a window of time in which a small batch of vertices and their edges are added. These edges connect the new vertices to each other and to the full graph generated from the last micro-batch. Second, the edges are equal in value i.e., the edges are not weighted or directed. We denote the graph from the previous iteration as $G_t = (V_t, E_t)$ where $V_t$ and $E_t$ are its sets of vertices and edges at time $t$ respectively. Let us refer to the vertices in the newly arriving batch as $V^*$ and the edges as $E^*$. We define a micro-batch from the stream of a node-grained dynamic graph $d$ as follows:

$$\delta^* = V^* \cup E^* \text{ where } \forall e_{j,k} \in E^* : v_j \in V^* \backslash V_t \lor v_k \in V^* \backslash V_t \quad (7)$$

Based on the cost of each step and the above-mentioned graph properties, we developed the IDWCC algorithm that works in three phases. First, it merges a micro-batch of the streaming data with the maintained evolving graph, updates the vertex statistics, and optimizes the graph. Second, it assigns the new vertices to the initial communities. Finally, it optimizes the WCC metric to generate better communities.

Table 1. Properties of the test graphs

| Data Sources | Vertices | Edges |
|---|---|---|
| **Amazon** | 334,863 | 925,872 |
| **DBLP** | 317,080 | 1,049,866 |
| **YouTube** | 1,134,890 | 2,987,624 |

IV. VALIDATION AND RESULTS

*A. Experimental Setup*

A distributed multi-cluster environment was used for our experiments applying Spark, Spark Streaming, and GraphX to implement the dynamic distributed graph. Eight identical machines were used, each having 8 cores 2.10 GHz Intel Xeon 64-bit CPU, 30 GB of RAM, and 300 GB of disk space to host and perform computations on the distributed graph data structure. We installed Apache Spark v2.2.0 on all the machines. Both DWCC and IDWCC algorithms were implemented using Scala 2.11[2].

---
[1] https://spark.apache.org/docs/latest/graphx-programming-guide.html#triangle-counting

*B. Data Source*

We used a set of different real-life undirected graphs that have ground-truth communities. We took these graphs from the SNAP data repository[3]. The selected graphs and some statistics about them are presented in Table 1. The use of multiple graphs for the experiments allowed us to compare the results for different graph sizes. In addition, it gave us the ability to experiment with different sizes of micro-batches easily. After loading a graph from the experimental sets, we ensured that it is clean and free from duplicates and self-edges. We also sorted the edges by the source vertices which made the graph canonical from the start and ready for processing.

*C. Complexity Analysis*

We compare the complexity of sequential implementation of our incremental algorithm to its static counterpart when both are applied to detect communities in a dynamic graph. Let $n$ be the number of vertices and $m$ the number of edges in the graph. We assume that the average degree of the graph $d$ at any point of time $t$ is $d = m/n$ and that real graphs (graphs are meant to be found in real-world with millions of vertices and edges) have a quasi-linear relation between vertices and edges $O(m) = O(n.logn)$. Under these assumptions, the complexity of the static WCC optimization algorithm, as calculated in Prat-Perez et al. [17] is $O(m.logn)$.

Now we calculate the complexity of a centralized version of the sequential incremental WCC optimization algorithm. For the first phase, we do not consider the complexity of merging the graphs since the operation is necessary for any dynamic graph. That leaves the cost of computing the triangles and degrees for the vertices in the new batch as follows:

$$O(m^*.d + m^*) = O(m^*.log\ n_{t+1})$$

The second phase requires sorting the vertices based on the local clustering coefficient. However, the vertices are already sorted from processing a previous micro-batch. Hence, the cost is only for organizing the new vertices in the right order which requires sorting the new vertices and then executing a full scan of the vertices in the worst-case $O(n^*+n_t) = O(n_t)$. For the third phase, let α be the number of iterations required to find the best possible communities, which is a constant. In each iteration, we compute in the worst-case $d+1$ movements for each vertex of type $WCC'_1$ which has a cost of $O(1)$. That makes the total cost as follows:

$$O(n.(d+1)) = O(m)$$

Next, we apply all the movements which are equal to the number of vertices, so it costs $O(n_{t+1})$. We also need to update, for each iteration in the second phase, the statistics $w$, $c_{out}$, $d_{in}$, and $d_{out}$ for each vertex and community, which has a cost of $O(m_{t+1})$. We sum all the costs to get the full cost of this phase $O(\alpha.(m_{t+1} + n_{t+1} + m_{t+1})) = O(m_{t+1})$. The full cost of the algorithm is the sum of the cost of the three phases: $O(m^*log\ n_{t+1} + n_t + m_{t+1})$. Since $m^* << n_{t+1}$, then $m^*.logn_{t+1} <<$ $n_{t+1}.logn_{t+1} < m_{t+1}$. The cost can be simplified to become $O(m_{t+1})$. This cost is much smaller than $O(m_{t+1}.logn_{t+1})$ the cost of applying the static algorithm on the whole graph during the $t$ th iteration.

*D. Experimental Details and Results*

We conducted the following three sets of experiments.

- Experiment *a*: We computed the efficiency of the optimizations used in IDWCC by comparing the execution time of each step of IDWCC with its counterpart in DWCC.
- Experiment *b*: We compared the quality of the results of DWCC, IDWCC, and SCD.
- Experiment *c*: Finally, we compared the quality of the results and the execution time of the DWCC and IDWCC algorithms for a dynamic distributed graph by executing the algorithms while updating the graph in real-time from a data stream. We executed 5 iterations of the DWCC and IDWCC algorithms based on the recommendations of Prat-Pérez et al. [19], the authors of the original WCC optimization algorithm.

We experimented with different graph sizes to compare the performances of the algorithms for different sizes of graphs and micro-batches of the streaming data. Initially, we constructed the graph from a bulk of static data already downloaded from the selected data sources. For the bulk data, the first two columns in Table 2 show the number of vertices and edges added to the graph respectively. Next, we appended new data to the existing distributed graph from the streaming data sources using micro-batches of data at a time. The number of edges added from the data streams and the corresponding numbers of micro-batches of data used for the updates are shown in columns 3 and 4 in Table 2 respectively. It can be noticed that we chose to use fewer micro-batches with the smaller size bulk graphs, and greater micro-batches for the larger size bulk graphs to limit the use of the resources on each iteration.

Table 2. Initial sizes and updated sizes of the test graphs

|  | Bulk Vertices | Bulk Edges | Stream Edges | # of Micro-batches |
|---|---|---|---|---|
| **Amazon** | 258,464 | 576,718 | 349,154 | 10 |
| **DBLP** | 253,119 | 852,754 | 197,112 | 10 |
| **YouTube** | 903,959 | 2666,836 | 320788 | 10 |
| **LiveJournal** | 768,792 | 13,997,342 | 20,683,847 | 30 |

Experiment *a* shows the benefits of the optimization techniques we applied to our IDWCC as shown in the algorithm listing. For both DWCC and IDWCC, we calculated the time that each step took to be executed on the full Amazon graph and aggregated them to compute the total execution time as shown in Fig. 2. The results show that the IDWCC has a shorter execution time compared to the DWCC algorithm. The *vertex statistics step* as defined in the implementation part in Section III is the one responsible for calculating each vertex triangle count and degree.

---
[2] https://github.com/TariqAbughofa/incremental_distributed_wcc
[3] http://snap.stanford.edu/

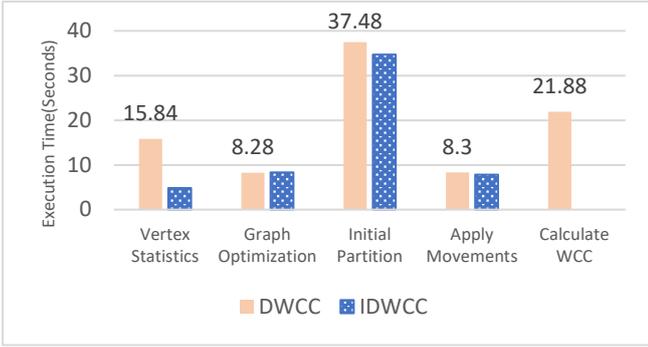

*Figure 2. Execution time for each step in DWCC vs IDWCC*

In Fig.2, we notice that the execution time was reduced in IDWCC by almost three times compared to DWCC as we counted the triangles and degrees only for the new vertices and the border vertices (vertices statistics Fig.2). The graph restructuring step had no change in execution time as no adjustments were made to the vertices. The iterative partitioning step had a small decrease in execution time caused by the way we altered this stage. The biggest gain was expected in memory consumption as proven later. Finally, the WCC optimization step, which calculates the WCC metric, an expensive operation, was eliminated completely (no column for IDWCC) from the IDWCC algorithm resulting in a great reduction in the computational cost.

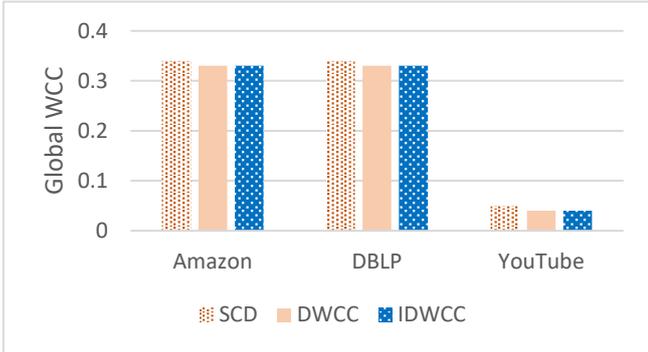

*Figure 3. Global WCC: SCD vs DWCC vs IDWCC*

Experiment *b* aimed at comparing the quality of results of the distributed graph community detection algorithms, DWCC and IDWCC. The existing implementation of the SDC algorithm already proved that it had better quality than many other centralized community detection algorithms while having faster execution [17]. These studies compared the quality of the results of the SDC to that of Louvain [33] and Infomap [27]. Therefore, instead of repeating similar experiments and comparing the effectiveness of WCC optimization to that of the other approaches such as Louvain and Infomap, we compared the quality of the communities produced by the SCD, DWCC, and IDWCC. We measure the quality of the communities by calculating the global WCC on the full test graph after appending data from the last micro-batch. The results are displayed in Fig. 3 which shows that both DWCC and IDWCC produced good WCC values. These WCC values show only up to 5% decrease from their SCD counterparts. On top of that, the IDWCC gives slightly better results than the DWCC algorithm. We do not show the results for the LiveJournal graph as both DWCC and IDWCC failed to process the whole stream. In both cases, the computational needs exceeded the available memory resources.

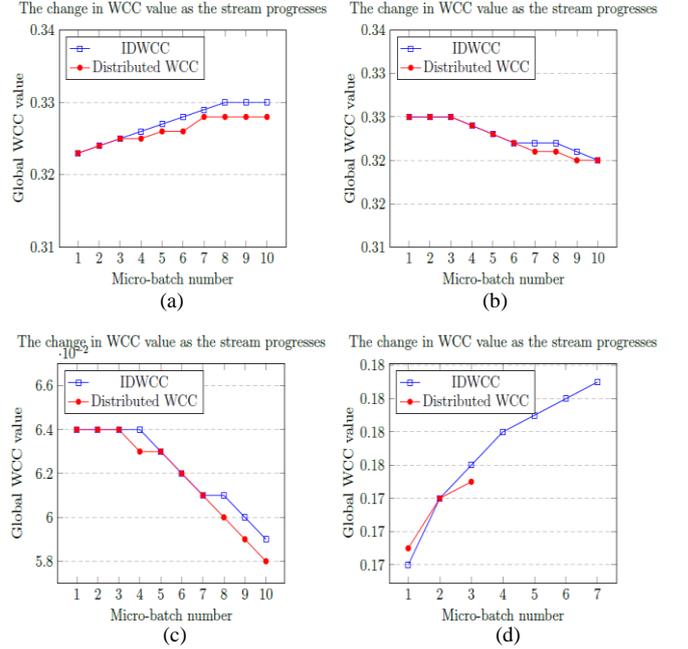

*Figure 4. Comparing WCC values (DWCC vs IDWCC for (a) Amazon; (b) DBLP; (c) YouTube; (d) LiveJournal*

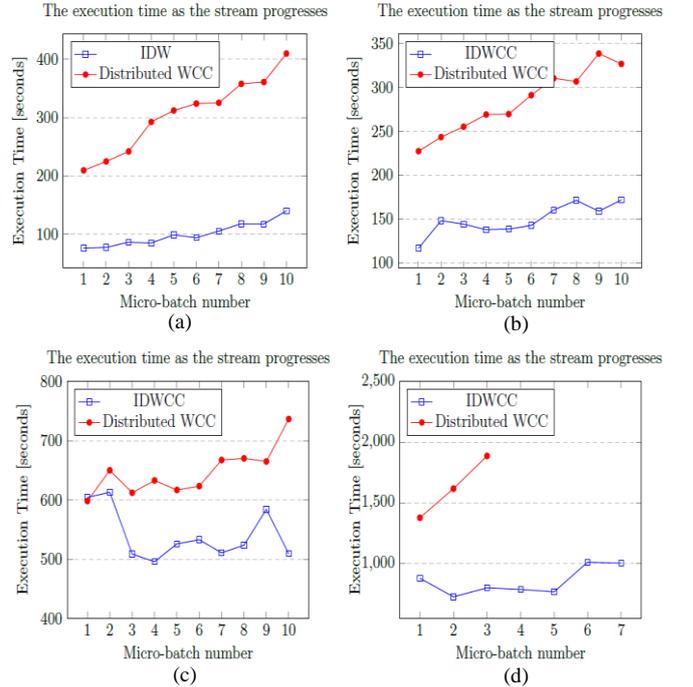

*Figure 5. Comparing WCC execution time (DWCC vs IDWCC for (a) Amazon; (b) DBLP; (c) YouTube; (d) LiveJournal*

Experiment *c* was designed to prove the quality of the results and the efficiency of the IDWCC algorithm compared to the DWCC algorithm. For each graph and each micro-batch in the stream as described in Table 2, we merged the micro-batch with the full graph. Then we applied both DWCC and IDWCC on the modified graph. In the streaming context,

DWCC finds the communities for the whole graph again, whereas our IDWCC finds communities only for the new vertices and reflects these changes on the old vertices.

For each micro-batch, we compared the global WCC values of the resulting graph generated by IDWCC and DWCC in addition to the execution times. Fig. 4 and 5 show the results of the experiments with streaming data. Missing data points indicate that the algorithm was not able to continue processing the graph due to insufficient memory problems (lack of resources). Fig. 4 clearly demonstrates that IDWCC produces communities with global WCC values that are very close to the ones produced by DWCC. We can even see that the results start to be better than DWCC in later iterations. Regarding the execution time, IDWCC performed two to three times better than DWCC. For the large LiveJournal graph, we see that both algorithms failed to continue with the available computational resources. However, IDWCC continued for 7 micro-batches before it crashed, while DWCC could only process up to 3 micro-batches. This shows that IDWCC has significantly less memory consumption than DWCC. The experiments show that our dynamic graph processing framework with IDWCC is capable of maintaining graphs up to 100 million edges while updating them with streaming data in under 50 seconds.

## V. CASE STUDY

We further examined the communities produced by IDWCC in a real-world application of dynamic graphs. As a case study, we chose the scenario of generating recommendations for product purchases using the Amazon streaming dataset [39]. The metadata for this graph is available on the SNAP website [38] and contains the titles of the products.

Table 3. Examples of communities produced by IDWCC on Amazon products data.

| Community # | Example 1 | Example 2 |
|---|---|---|
| community #1 | Gulliver's Travels | Robinson Crusoe: His Life and Strange |
| | Science Fiction Classics | Surprising Adventures |
| | of H.G. Wells | Treasure Island |
| | Swiss Family Robinson | Gulliver's Travels |
| | The War of the Worlds | The Swiss Family Robinson |
| | Anne of Avonlea | Robinson Crusoe: Life and Strange |
| | | Surprising Adventures |
| community #2 | Merchant of Venice | Hamlet: The New Variorum Edition |
| | The Merchant of Venice | Hamlet |
| | Macbeth | The Merchant of Venice |
| | Othello: The Applause | A Midsummer Night's Dream |
| | Shakespeare Library | Othello |
| | Much Ado About | |
| | Nothing | |
| community #3 | 1984 | To Kill a Mockingbird |
| | A Separate Peace | John Knowles's a Separate Peace |
| | Lord of the Flies | Joseph Heller's Catch-22 |
| | Romeo and Juliet | The Grapes of Wrath |
| | 1984 | 1984 |

The product recommendation problem aims to find products that are usually bought together to suggest them to the users. This graph represents a network of products. Each vertex represents a product, and each edge connects two products if they are frequently purchased together. Therefore, the communities in the graph constructed using the Amazon data would represent similar products that are frequently purchased together and can be used to provide recommendations about products to the customer. We ran the IDWCC algorithm on the graph and chose three communities which are reported in Table 3. Table 3 has 10 randomly selected vertices from each community. In the case of the first community, we see that it is formed of classic novels. The second community consists of Shakespearean literature. Finally, the third one is mostly political and allegorical novels. We observe that the algorithm can perform a good selection of the relations in the graphs to give meaningful communities.

## VI. CONCLUSION

Detecting communities in very large graphs offers computational challenges and existing algorithms do not offer a feasible solution for large dynamic graphs [7]. In this paper, we propose the IDWCC algorithm as a solution to the community detection problem for large dynamic distributed graphs that need to be updated continuously from streaming data. We demonstrate the efficacy of our solution by implementing a prototype using cutting edge Spark, Spark Streaming and GraphX to create and maintain a large in-memory distributed graph over a multi-cluster distributed infrastructure. We begin by conducting a study on the use of the Weighted Community Clustering (WCC) metric to detect communities with Apache Spark. Next, we present the implementation of a Distributed WCC (DWCC) optimization using Apache Spark and GraphX to detect communities in static graphs. Finally, we propose and demonstrate a novel Incremental Distributed WCC (IDWCC) algorithm for detecting communities in large node-grained dynamic distributed graphs. IDWCC improves the DWCC optimization by assigning the newly added vertices to the most suitable communities in a distributed graph and by optimizing some of the computational steps. The algorithm is implemented in Scala using the in-memory GraphX structure and is executed on a distributed multi-cluster environment using Apache Spark. The experiments showed that IDWCC outperforms DWCC for large dynamic graphs. IDWCC produced the same or better WCC values compared to DWCC. It was also two to three times faster than DWCC. The memory consumption was more optimized in IDWCC as well. To the best of our knowledge, IDWCC is the best performing incremental community detection algorithm for node-grained dynamic distributed graphs. We also demonstrated and validated the usability of dynamic community detection using IDWCC for a real-life e-commerce use case scenario of product recommendations using Amazon product data.

As future work, we like to study the stability of the quality of IDWCC over a long period of time with a goal to assess the need of applying the DWCC optimization periodically on the full graph to maintain high accuracy of the results in the case of result degradation. We also aim to address the memory consumption problem of the IDWCC algorithm which causes a bottleneck when computing new communities for each vertex. We plan to further optimize the iteration phase by limiting the number of vertices for which we update the communities in each iteration. This may be done by using statistics calculated from the previous iterations. We only addressed undirected unweighted node-grained dynamic graphs in this research. We would like to extend our framework to work with edge-grained dynamic graphs and explore the effect of the edge weights in the community detection process.

ACKNOWLEDGMENT

The authors wish to thank the Queen's University Centre for Advanced Computing (CAC) for providing access to computing resources to run our experiments.